# The State-of-the-Art in Air Pollution Monitoring and Forecasting Systems using IoT, Big Data, and Machine Learning

**Amisha Gangwar**[1,2], **Sudhakar Singh**[1] *****, **Richa Mishra**[1], **Shiv Prakash**[1]

[1] Department of Electronics and Communication, University of Allahabad, Prayagraj, India
[2] School of Information Studies, Syracuse University, New York, US

*Corresponding author
E-mail addresses:
agangwar@syr.edu (Amisha Gangwar)
sudhakar@allduniv.ac.in (Sudhakar Singh)
richa_mishra@allduniv.ac.in (Richa Mishra)
shivprakash.cse@gmail.com (Shiv Prakash)

**Abstract**

The quality of air is closely linked with the life quality of humans, plantations, and wildlife. It needs to be monitored and preserved continuously. Transportations, industries, construction sites, generators, fireworks, and waste burning have a major percentage in degrading the air quality. These sources are required to be used in a safe and controlled manner. Using traditional laboratory analysis or installing bulk and expensive models every few miles is no longer efficient. Smart devices are needed for collecting and analyzing air data. The quality of air depends on various factors, including location, traffic, and time. Recent researches are using machine learning algorithms, big data technologies, and the Internet of Things to propose a stable and efficient model for the stated purpose. This review paper focuses on studying and compiling recent research in this field and emphasizes the Data sources, Monitoring, and Forecasting models. The main objective of this paper is to provide the astuteness of the researches happening to improve the various aspects of air polluting models. Further, it casts light on the various research issues and challenges also.

## 1. INTRODUCTION

Air pollution is exacerbating life quality on the earth. This alarming situation needs proper solutions to monitor the air, locate the sources causing air pollution and limit the use of those sources. Outside from the parking lots to inside the houses, the source of air pollution is present everywhere. Nearly 4 million people die per year due to illness caused by inefficient techniques used for household cooking. Among these deaths, 27% are from pneumonia, the other 27% from ischaemic heart disease, 20% from COPD (chronic obstructive pulmonary disease), remaining 18% and 8% are stroke and lung cancer, respectively [1]. Household Air Quality (HAQ) depends mainly on the kind of cookstove used and the geometry of the kitchen. Classic traditional cookstoves must be replaced with clean cooking solutions like Induction stoves, Forced drafted cookstoves, and Traditional cookstoves with hood structures. These cooking solutions lower the concentrations of Black Carbon, Carbon Monoxide (CO), and other pollutants emitted during the cooking process [2]. This all is not just limited to household pollution. Ambient pollution has more worsen outcomes. Smog, the brownish-yellow coloration in the atmosphere, is caused due to the presence of pollutants like ozone, nitrogen oxides, and sulfur oxides. When nitrogen oxides released from automobiles, factories, or power plants react with any volatile organic compound in the presence of sunlight results in photochemical smog in the atmosphere, the end products of these reactions are severely harmful to human health [3]. Analysis of health impacts in crowded urban areas of the Bangkok Metropolitan Region (BMR) was done through photochemical smog modeling of $PM_{2.5}$, and studies showed that meeting the limits set by World Health Organisation (WHO) for $PM_{2.5}$ can avoid 1,415 death annually in that region [4]. According to Iaccarino et al. [5], there is a strong connection between ambient pollution and loss of cognitive abilities in older adults, whereas continuous exposures to Carbon Monoxide coming from fuel-burning can lead shortly to fatigue, drowsiness, weakness, and difficulty in breathing and later severely to diseases like lung cancer and brain cancer. Haemoglobin has more affinity to Carbon Monoxide than Oxygen. If a person gets exposed to Carbon Monoxide, blood haemoglobin will make a strong bond with it and hinder the supply of oxygen to other parts [6]. The presence of sulfur dioxides and nitrogen oxides after the combustion of fossil fuels causes acid rain during precipitation. This acid rain disrupts humans and even wildfire and vegetation [7]. Juzhang Ren [8] studies show that ozone pollution near winter wheat fields also affects the economic yield by a 2.8% loss. All these damages need to be appropriately scaled. Hence, an interesting value evaluation system was established by Dongyue Liu [9], which assesses the environmental damages due to air pollution and provides legal compensation for the individual or organization that caused the one. However, merely evaluation methods are insufficient to address these issues; an active mechanism is required to cease or control the pollution from all the primary sources. For example, a setup can be constructed to regulate the pollution within defined standards from industries, or these systems can be installed around the pulmonology department of the hospital. From a commercial standpoint, these systems come with great investment results as these systems can be used by every sector dealing with bad air quality.

To preserve the air quality, various researches are going around the problem, i.e., pollution detection, Monitoring, source detection, and forecasting. These help in planning calculative strategies towards the problem [10]. For example, in [11], a low-cost setup was proposed using the Internet of Things (IoT) for monitoring purposes, whereas a similar approach was also made in [12] to monitor the data and further extended to a routing and predicting system. Studies mentioned in [13]–[15] are extended to examine what factors affect the problem more closely, what algorithms may improvise the forecasting feature, how to place the IoT setups, and how stable and economical the models are in real life. So, there are many aspects of the problem that need to be considered for a better solution.

Advanced technologies are making their imprints in every field, providing a whole new dimension to every other area. Since the last few decades, Artificial Intelligence (AI), Machine Learning (ML) algorithms, IoT, and Big Data tools are overtaking all primitive methods, giving out cost-friendly and many promising results for the stated problem [16], [17]. Fig. 1 shows how these advanced technologies are used in Monitoring and prediction system. Also, Fig. 2 illustrates a general model that should be installed at every air station to take calculative measures in that area to deplete the air pollution in that area. In this direction, the main objective of this paper is to provide an overview of all the comprehensive research and an insightful review of the current state-of-the-art regarding the problem and the significant issues and challenges for an effective and efficient system. The main contributions of this paper can be summarized as follows.

- Various criteria and standards set by different governments and agencies for Air Quality Index are investigated.
- A systematic review of IoT-based real-time air monitoring systems is carried out, and optimal deployment of sensors is also emphasized.
- A comparative analysis of machine learning over classical statistical forecasting methods has been done.
- A systematic review and critical analysis of big data and machine learning-based models and comparisons on the various parameters are carried out.
- Recent research issues and challenges for efficient air pollution monitoring and forecasting are highlighted.

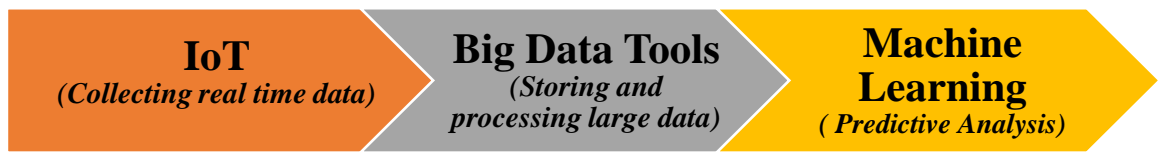

Fig. 1: Advanced Technologies used in recent Predicting and Monitoring Models.

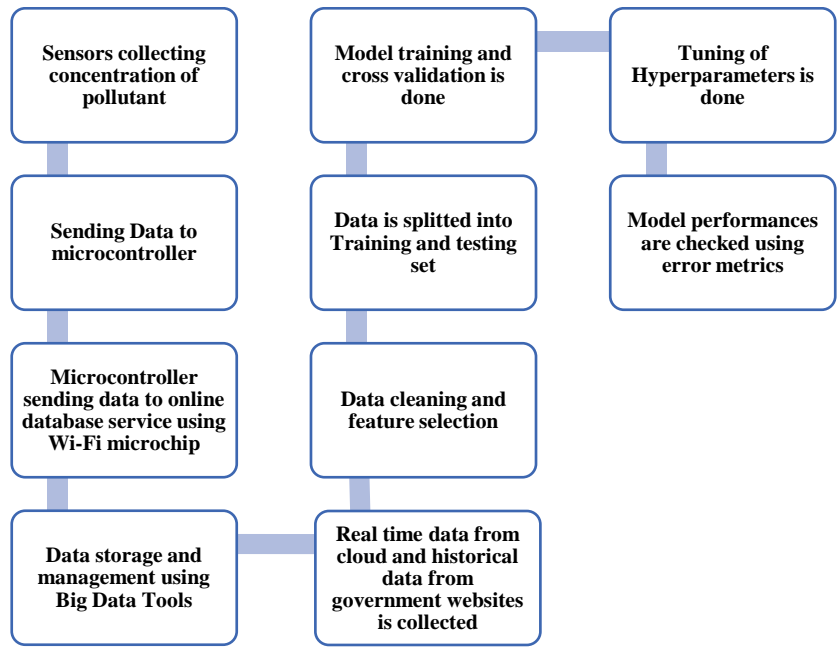

Fig. 2: IoT, Big Data, and ML-based Air Pollution Monitoring and Prediction System.

This is a review paper for air pollution monitoring and forecasting techniques. It is initiated with Section 1, which is a general introduction about air pollution and its effects along with the objective of this study. Section 2 discusses various air pollution standards proposed by the authorities and researchers. Section 3 presents the IoT infrastructure, reviews the various IoT-based models, and discusses the optimal placement of IoT setups. Section 4 discusses machine learning over classical statistical forecasting methods, examines the different Big Data and ML-based models, and compares them on multiple parameters. Section 5 focuses on the research issues and challenges for an effective air pollution monitoring and forecasting system, and Section 6 concludes the paper with the direction of future works. Fig. 3 schematically represents the structure of the paper after the introduction.

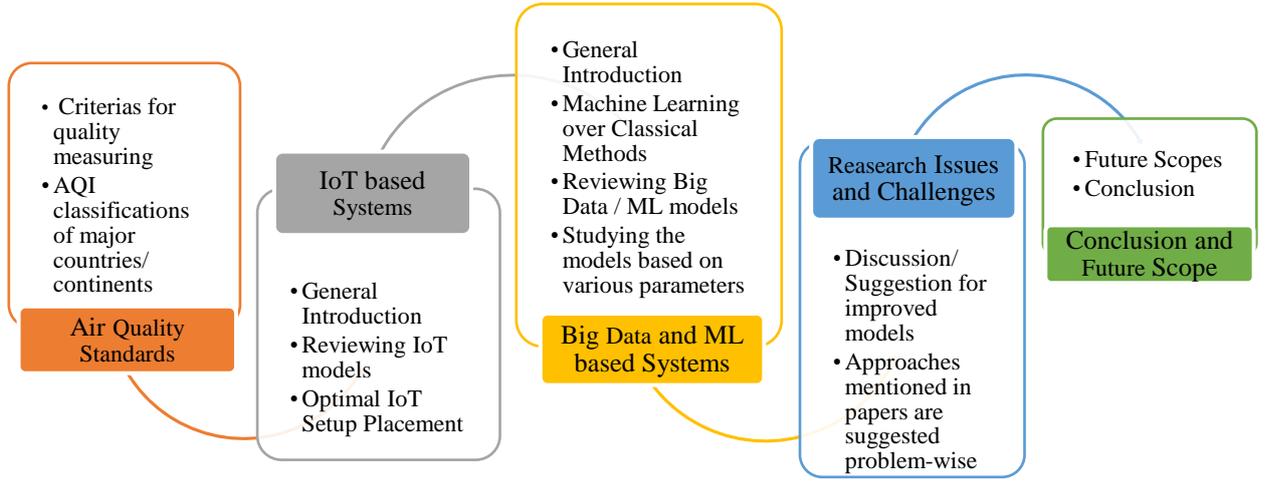

Fig. 3: Structure of the Paper

## 2. AIR QUALITY STANDARDS

Air pollution is one of the sustainable concerns that is also mentioned in Sustainable Development Goals [18]. It is explicitly mentioned in SDG 3.9, SDG 11.6 and indirectly linked with SDG targets [19]. Unsustainable growth in urbanization as well as in industrialization leads to major respiratory illness. The various governments and agencies have set the air quality standards, which are described as follows.

### 2.1 US Air Monitoring Criteria

Six pollutants are declared as criteria pollutants by U.S. EPA (Environmental Protection Agency)– Carbon Monoxide (CO), Lead (Pb), Nitrogen Dioxide ($NO_2$), Ozone ($O_3$), Particulate Matter (PM) or Particle Pollution, and Sulphur Dioxide ($SO_2$). There are two subcategories for **Particulate Matter** based on pollutant size – $PM_{2.5}$ and $PM_{10}$. EPA has also fixed the averaging time and levels for each of the aforementioned pollutants, as shown in Table 1. Primary and secondary, mentioned in Table 1, are the state in which commodities are effective. Primary mainly includes the health of senior citizens, infants, asthma patients, whereas Secondary consists of the health of crops, animals, vegetation. The units used for levels are ppm – parts per million by volume, ppb – parts per billion by volume, μg/m³ – micrograms per cubic meter of air [20]. The Air Quality Index (AQI) is a measure of how polluted the air is. It is a piecewise linear function of air pollutant concentration. As described in [21] and [22] , the following formula is used to calculate AQI.

$$AQI = \frac{AQI_{high} - AQI_{low}}{PC_{high} - PC_{low}} (PC - PC_{low}) + AQI_{low} \qquad (1)$$

where,
$AQI$ = Air Quality Index,
$PC$ = pollutant concentration,
$PC_{low}$ = concentration-breakpoint ≤ $PC$,
$PC_{high}$ = concentration-breakpoint ≥ $PC$,
$AQI_{low}$ = index-breakpoint respective to $PC_{low}$,
$AQI_{high}$ = index-breakpoint respective to $PC_{high}$.

Table 1: Criteria Pollutants to Measure Pollution [20]

| Pollutant | Primary (P)/Secondary (S) | | Averaging Time | Level |
|---|---|---|---|---|
| Carbon Monoxide (CO) | P | | 8 hours | 9 ppm |
| | | | 1 hour | 35 ppm |
| Lead (Pb) | P and S | | Rolling 3-month average | 0.15 μg/m$^3$ |
| Nitrogen Dioxide (NO$_2$) | P | | 1 hour | 100 ppb |
| | P and S | | One year | 53 ppb |
| Ozone (O$_3$) | P and S | | 8 hours | 0.070 ppm |
| Particle Pollution (PM) | PM$_{2.5}$ | P | 1 year | 12.0 μg/m$^3$ |
| | | S | 1 year | 15.0 μg/m$^3$ |
| | | P and S | 24 hours | 35 μg/m$^3$ |
| | PM$_{10}$ | P and S | 24 hours | 150 μg/m$^3$ |
| Sulphur Dioxide (SO$_2$) | P | | 1 hour | 75 ppb |
| | S | | 3 hours | 0.5 ppm |

After calculating the AQI value, it is referred to Table 2 to check the classification. Each country has its air quality indices classification. Table 2 is for the United States [23].

Table 2: AQI Classification [23]

| Range of AQI Values | Levels of Health Concern |
|---|---|
| 0 – 50 | Good |
| 51 – 100 | Moderate |
| 101 – 150 | Unhealthy for Sensitive Groups |
| 151 – 200 | Unhealthy |
| 201 – 300 | Very Unhealthy |
| 301 – 500 | Hazardous |

## 2.2 European Air Monitoring Criteria

European Union (EU) has a whole extensive body of legislatures that monitor ambient air pollution. They have set the standards as well as the objective for each pollutant present in the air [24]. European Environment Agency in November 2017 declared the EAQI (European Air Quality Index), and since then, it is being encouraged [25]. Selected EU standards are summarised in Table 3, taken from [25].

Table 3: EU Standards [25]

| Qualitative name | Index or sub-index | Pollutant (hourly) concentration in µg/m³ | | | |
|---|---|---|---|---|---|
| | | $NO_2$ | $PM_{10}$ | $O_3$ | $PM_{2.5}$ (optional) |
| Very low | 0–25 | 0–50 | 0–25 | 0–60 | 0–15 |
| Low | 25–50 | 50–100 | 25–50 | 60–120 | 15–30 |
| Medium | 50–75 | 100–200 | 50–90 | 120–180 | 30–55 |
| High | 75–100 | 200–400 | 90–180 | 180–240 | 55–110 |
| Very high | >100 | >400 | >180 | >240 | >110 |

## 2.3 Indian Air Monitoring Criteria

Out of the world's 20 most polluted cities, 13 cities are from India alone. Some of them are Kanpur, Faridabad, Gaya, Varanasi, Patna, Delhi, and Lucknow [26]. This poor situation of breathing environment needs the attention of Indian government authorities and the participation of citizens. Under Swachh Bharat Abhiyan in October 2014, the government announced National Air Quality Index [27]. Along with State and Central Pollution Control Boards, National Air Monitoring Programs (NAMP) operate in 300+ cities and have 700+ monitoring stations [28]. Table 4 shows the Indian Air Quality Index [29].

Table 4: National Air Quality Index (NAQI) of India [29]

| AQI Category (Range) | $PM_{10}$ (24hr) | $PM_{2.5}$ (24hr) | $NO_2$ (24hr) | $O_3$ (8hr) | CO (8hr) | $SO_2$ (24hr) | $NH_3$ (24hr) | Pb (24hr) |
|---|---|---|---|---|---|---|---|---|
| **Good (0–50)** | 0-50 | 0-30 | 0-40 | 0-50 | 0-1.0 | 0-40 | 0-200 | 0-0.5 |
| **Satisfactory (51–100)** | 51-100 | 31-60 | 41-80 | 51-100 | 1.1-2.0 | 41-80 | 201-400 | 0.5-1.0 |
| **Moderately polluted (101–200)** | 101-250 | 61-90 | 81-180 | 101-168 | 2.1-10 | 81-380 | 401-800 | 1.1-2.0 |
| **Poor (201–300)** | 251-350 | 91-120 | 181-280 | 169-208 | 10-17 | 381-800 | 801-1200 | 2.1-3.0 |
| **Very poor (301–400)** | 351-430 | 121-250 | 281-400 | 209-748 | 17-34 | 801-1600 | 1200-1800 | 3.1-3.5 |
| **Severe (401–500)** | 430+ | 250+ | 400+ | 748+ | 34+ | 1600+ | 1800+ | 3.5+ |

These criteria and standards are followed in most of the literature. For examples, Kok et al. [30] and Alaoui et al. [31] used criteria similar to the US, whereas Nashh et al. [32] mentioned the US and

European Standards. European Standards is also used in Lazrak et al. [33]. Indian Standards are used in Moses et al. [12] Monitoring and Re-routing system.

## 3. INTERNET OF THINGS BASED MODELS

### 3.1. Internet of Things (IoT)

It is a collection of smart embedded sensors, devices, and software that helps to collect data and exchange it over the internet [34]. It provides the system with real-time data. The IoT architecture consists of the perception layer, network layer, and third-party application or cloud layer [35].

**i) Perception Layer**

This layer is also called the sensing layer. It consists of sensors. Sensors sense the data and send it periodically to the gateway or the cloud [35]. So, these sensors measure the levels of the specific pollutant. There are particular kinds of sensors to sense the particular pollutant, shown in Fig. 4.

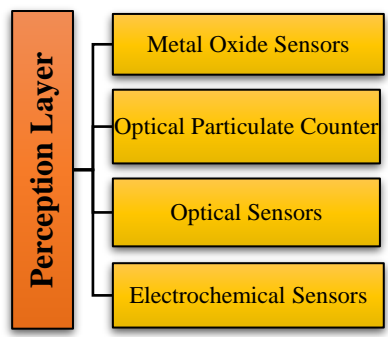

Fig. 4: Various kinds of Sensors

*a) Metal Oxide Sensors*
These are low-cost sensors with good sensitivity and long response times. These sensors are easily affected by temperature and humidity. It is usually used to measure $O_3$, CO, and $NO_2$ [36].

*b) Optical Particulate Counter*
These are moderate-cost sensors with a fast response time and a sensitivity range of 1 µg/m³. It can measure the size of the particle. Hence, this is used for measuring $PM_{2.5}$ and $PM_{10}$, and the output depends on various factors such as color and density, humidity, refractive index, and shape [36].

*c) Optical Sensors*
These are moderate-cost sensors used to measure $CO_2$ and CO. They have good sensitivity for $CO_2$ with a response time of 20-120 seconds [36].

*d) Electrochemical Sensors*
These sensors are of moderate cost with good sensitivity (mg/m$^3$ to µg/m$^3$) and quick response time. It is usually used for measuring $NO_2$, $SO_2$, $O_3$, NO, and CO [36].

Table 5 gives the names of some specific IoT sensors and their details like cost (in Indian rupees), usages, and range of detection. The costs mentioned here are taken form e-commerce portals.

Table 5: Names and Attributes of few Sensors

| Attributes / Sensors | Cost | Used for | Range of detection | Ref. |
|---|---|---|---|---|
| MQ9 | ₹ 100 – 300 | CO/Combustible Gas | 200~10000 ppm | [37] |
| MQ131 | ₹ 1500 – 2500 | $O_3$/Ozone | 10~1000 ppm | [38] |
| MQ135 | ₹ 150 – 500 | $NH_3$, $NO_x$, Alcohol, Smoke | 10~300 ppm ($NH_3$/Alcohol) 10~1000 ppm (Benzene) | [39] |
| MQ7 | ₹ 170 – 300 | CO | 20~2000 ppm | [40] |
| DHT11 | ₹ 150 – 5000 | Temperature & Humidity | 0~50ºC (Temperature) 20% ~ 80% (Humidity) | [41] |

### ii) Network Layer

This layer is the bridge between perception and application layers. This layer consists of gateways. Gateways are local schedulers, regulators, and processors. They do perform light processing to decrease unnecessary transmission [35].

### iii) Third-Party Application or Cloud Layer

This layer provides the user with an interface to work on the data. It stores and analyzes the data and provides the final result. It can also perform real-time processing and learn through many available ML algorithms [35]. Fig. 5 represents the flow of collected data through all layers.

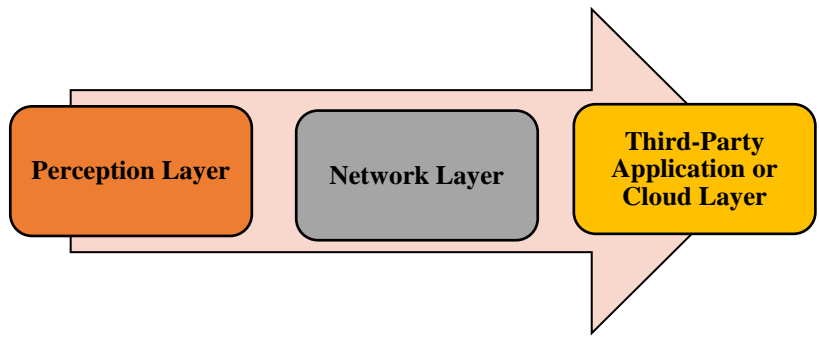

Fig. 5: Flow of data in an IoT setup

## 3.2 IoT Based Air Monitoring Systems

The IoT-based air monitoring system provides real-time location data for mining purposes. To construct an IoT setup, various embedded sensors, smart objects, network/ internet, software, cloud services, and devices are structured and integrated to meet the stated goal. Fig. 6 depicts a general design of IoT based Air Monitoring Systems.

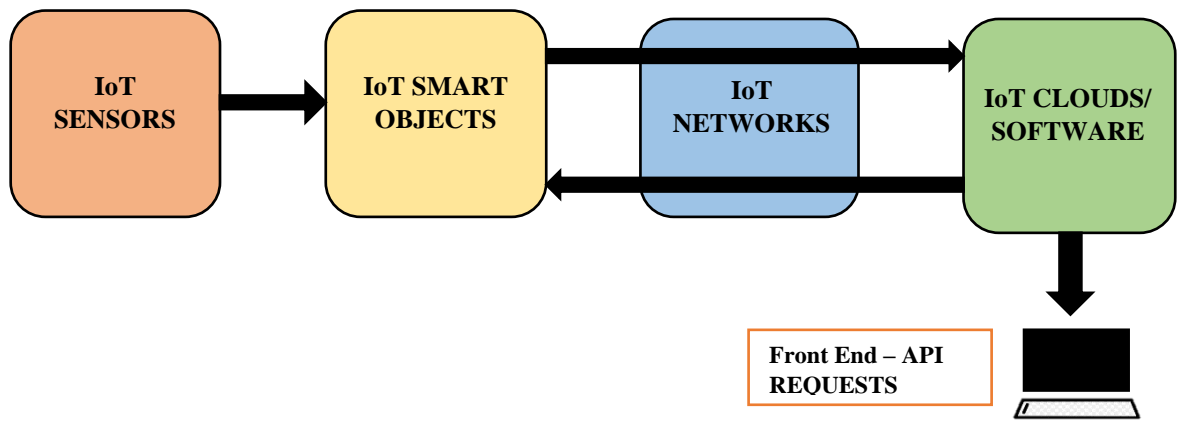

Fig. 6: Working of IoT based Models

In Fig. 6, all the setup components are connected via Network/Internet providing communicative abilities to the model. IoT Sensors (like MQ9, MQ131, MQ135, MQ7, DHT11) can be installed to collect the features of air and send it to IoT Smart Objects (like Wi-Fi Modular, Raspberry Pi, Arduino Microntrolllers). These Smart Objects then transfer the data to software or cloud using the IoT Networks (like Local and Personal Area Networks (LAN/PAN), Cellular, Mesh, Low Power Wide Area Networks (LPWAN)). These data can be sent to the front-end using API (Application Programming Interface) Integration. Further, it can be used for analysis and forecasting [42].

Borges et al. in [11] used the LM317 integrated circuit (positive voltage regulator), MQ7 sensors to measure CO gas, an ESP8266, which has the advantage of transferring data wirelessly by using TCP/IP protocol. With ESP8266, ESP8266-12E is also presented in their setup, which was assisted using Arduino sketch via USB (Universal Serial Bus).

Ayele et al. in [15] have used a DHT11 sensor for real-time humidity and temperature data. The MQ135 sensor is used for generating gas data, and both sensors are connected with the ESP8266 microcontroller to send the data to the webserver. Whereas, Kiruthika et al. [43] used Raspberry Pi as the mainboard, which is interfaced with the MQ5 gas sensor and time-humidity sensor, and whole data is transmitted via ESP8266 to the webserver. The python script is used to code. If humidity, temperature, or gas value crosses the threshold values, the actuators are turned ON. If it is below the threshold value, then it is stored in the MySQL database and queried via ThingSpeak. This setup is a truly low-cost, effective air monitoring system.

In [12], Moses et al. used various sensors, MQ131 for Ozone, MQ7 for CO, 110-602 Sulphur dioxide electrochemical sensor, $NO_2$, and $PM_{2.5}$ sensors. These all are connected to NB-IoT (NarrowBand-Internet of Things) and Raspberry Pi. NB-IoT collects the data for the sensor and sends it to the network via a narrow band device, and by installing MQTT protocols, the NB-IoT can directly send data to the cloud service. After this, the data is used to calculate AQI, and the estimated AQI is updated on Google Maps using an API. The authors have considered the maximum number of pollutants for monitoring the air.

Srivastava et al. [44] used HPMA115S0 sensor for $PM_{2.5}$ and $PM_{10}$, SHT10 for temperature, humidity, and CO, and Raspberry Pi 3 Model B as the microcontroller. Data collected is stored on the server, and as soon as any of the pollutants exceeds its threshold values as declared by Central Pollution Control Board (CPCB), it sends an emergency notification on the android application.

Okokpujie et al. [45] used an MQ135 sensor connected through Arduino to ESP 8266 Wi-Fi Module. A 16 by 2 LCD Screen is used to display the data locally. Data was sent to the ThingSpeak server, and graphs were plotted to study the air quality. In [46], Gupta et al. proposed an air monitoring system using IoT hardware, ThingSpeak, and android application. DHT11, MQ-2, and SDS021 sensors were used to collect the pollutant values (i.e., Temperature, Humidity, CO, Smoke, LPG, $PM_{2.5,}$ and $PM_{10}$). These sensors are then connected to Raspberry pi, and an Analog-Digital Converter is also connected for converting analog signals to digital ones. These data are sent and stored on ThingSpeak via API on the private channel. On ThingSpeak, the data is plotted against the date and time. This data on ThingSpeak will create three JSON files which will later be fetched by android studio by JavaScript Object Notation (JSON) parsing. Firebase API is then included in Android or iOS applications to get the real-time database, storage, analytics, and other data. Users can access pollutants concentration for any date and time.

An interesting model was also shown by Esfahan et al. [47], who proposed an IoT-based indoor air pollution monitoring system. Based on the US Environmental Protection Agency (US EPA) guideline, the indoor air quality index (IAQI) was taken into account. Sensors were connected to ESP32 based microcontroller, which transfers the data to the server using Wi-Fi. A fan is also connected to cool the setup if needed. The data collected is sent to the third-party application Blynk on smartphones. Then two experiments were carried out; the first one was carried out in computer laboratories of the School of Engineering, University of Warwick. The result with 100 occupants showed a poor level of $CO_2$. The second experiment was done in a kitchen where food was being prepared with an occupant. $CO_2$ levels were high, whereas $PM_{2.5}$ and TVOC (Total Volatile Organic Compounds) levels were also high but were at an acceptable limit.

### 3.3  Optimal IoT Setup Placement Studies

The optimal placement of IoT setup means how a proper IoT network can be developed across the targeted area to get more precise results. It is an essential aspect of effective and efficient air monitoring systems. In this direction, Sun et al. [13] proposed and optimized the sensor placement strategies in Cambridge city, U.K. The objective of this study includes the Monitoring of traffic emissions, protection to vulnerable sections of citizens, and maximizing the satisfaction of citizens. Dataset of this study comprises Traffic pattern data, population data, and data where vulnerable people spend their most time. The authors suggested reorienting the present sensor placement pattern in Cambridge and concluded the study on two points: first, to increase the number of low-cost air pollution setups, and second, the sensor setups must be spread uniformly across the city to fulfill the objectives. Optimal sensor placement is a fundamental and widespread optimization-based problem. Many authors have investigated the optimal placement of sensors. For example, the studies on the Gaussian Process model for optimal sensor placement by Dur et al. [48], Krause et al. [49], Longi et al. [50], and [51].

### 4.  BIG DATA AND MACHINE LEARNING-BASED MODELS

### 4.1.  Big Data

Big data is not simply the data big in size that requires big storage; in fact, it is one of its major properties. It is a complex, unstructured, rapidly generating data huge in volume [52]. Hrehova [53] has presented some definitions of big data from various sources. Generally, big data is defined in terms of its V's concepts. Initially, it was defined with three V's concepts: Volume, Velocity, and Variety by Doug Laney [54], and then expanded to include fourth V: Veracity [55]. At present, it includes more than twenty V's [56]. Big data imposed new challenges like the scalability and efficiency of the traditional ML and data analytics algorithms for big data. Big data analytics provides the solution by implementing these traditional algorithms on various big data platforms [57].

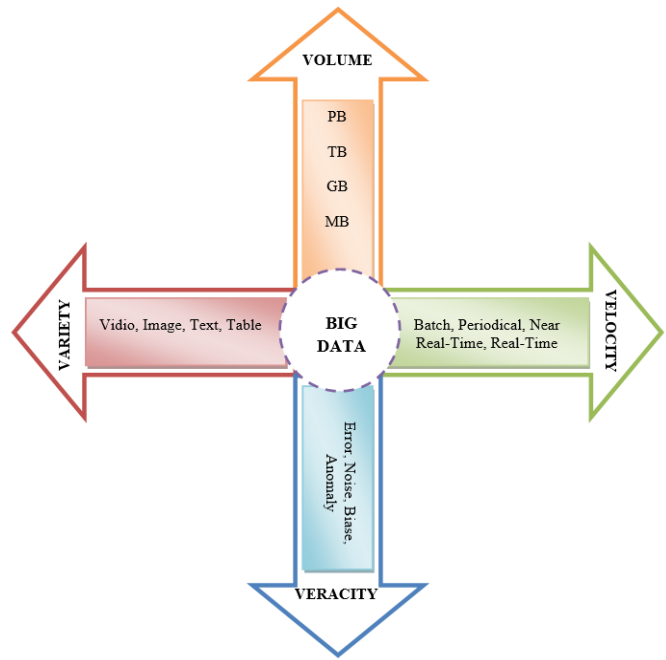

Fig.7: 4 V′s Concept of Big Data

Fig. 7 describes the 4 V's definition of Big Data. Volume refers to data size, Variety means the kind of data generated (structured, semi-structured, unstructured), Velocity is the speed of generated data, and Veracity is the grade until data can be trusted.

## 4.2. Machine Learning (ML)

ML is the process of training the computer so that it can automatically improve its performance over time without being explicitly programmed for a specific task. The machine learns using various algorithms present, and then it is trained on the particular amount of data known as train data, and later its performance is evaluated on an entirely new set of data that is not encountered before, known as test data. Some applications of machine learning are pattern recognition, image analysis, predictions, recommendation, etc. [58]. ML can broadly be divided into three categories which are shown in Fig. 8.

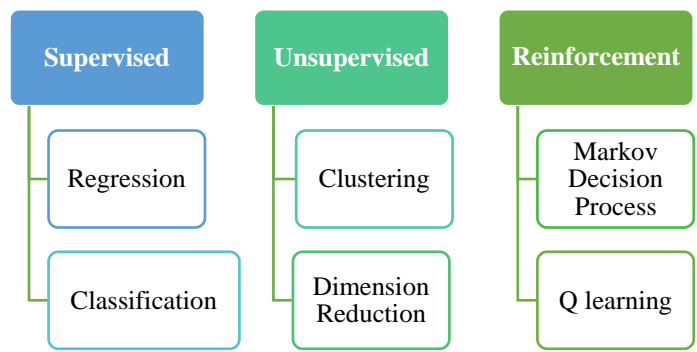

Fig. 8: ML Algorithm Classification

### i) Supervised Learning

In Supervised Learning, labeled data are used for training purposes. Supervised learning algorithm includes Regression and Classification algorithms. Regression algorithms are mainly used for predicting an entity and determining the relation between quantitative data. It has explicitly Linear Regression, Decision Trees, Bayesian Networks, and Fuzzy Classification. Classification algorithms are used for classifying the set of data into new classes, and it includes Logistic Regression, Classification Tree, Random Forest, and Support Vector Machines (SVM) [58], [59].

### ii) Unsupervised Learning

It is just the opposite of Supervised Learning. No labeled or solved data is present in the training set. The machine has to find out its solution. It mainly includes Dimension Reduction and Clustering algorithms. Dimension Reduction algorithms reduce the large dataset with many features into data with fewer features while maintaining the fundamental aspect. Principal Component Analysis (PCA), Tensor Reduction, Random Projection, and Multidimensional Statistics are included in Dimension Reduction. Whereas Clustering algorithms partition the input data into clusters based on certain criteria. It includes Gaussian Mixture Models, Genetic Algorithms, Hierarchical Clustering, and K-means Clustering [58], [59].

### iii) Reinforcement Learning

In Reinforcement learning, the agent learns to make decisions over time by consequences. It uses trial & error using feedback methods to learn. Like Supervised Learning, it also uses the mapping between input and output but uses punishment and rewards as feedback. The two important models for reinforcement learning are Markov Decision Process and Q Learning. The applications of Reinforcement learning are self-driving cars, gaming, recommendation system, financing, trading, etc. [60], [61].

### 4.3 Machine Learning over Classical Statistical Forecasting Methods

Classical statistical forecasting methods are used for univariate time series problems. Some of the classical statistical methods for time series are as follows and the same is summarized in Fig. 9.

i. **Naïve 2:** This method sets the forecast to last observation is usually helpful if the data consists of long periods of apparent ups and downs. Hence, it is also known as the adjusted random walk model for the season [62].
ii. **Simple Exponential Smoothing (SES):** This method is suitable for the dataset with no clear seasonality or trends [62].
iii. **Holt:** It is an extension of SES for the data with trends, and to capture the seasonality, Holt-Winter method is used [62].
iv. **Damped Exponential Smoothing (DES):** Previous methods assume that trend would go on forever, and to eradicate this idea, Damped exponential smoothing is done [62].
v. **Theta Method:** This method decomposes the original data into seasonality-based theta lines [63].
vi. **ARIMA:** ARIMA stands for the Auto-Regressive Integrated Moving Average and aims to correlate features. It is a widely used approach for time series problems [62].

vii. **ETS:** ETS stands for Error, Trend, and Seasonality. It is an exponential smoothing model in which decomposition plots help to find out whether to add, multiply, or leave out these trends, errors, and seasonality [62].

viii. **HMM:** HMM stands for Hidden Markov Model**,** a statistical Markov model that detects the hidden states to learn about Markov Chains. It is a probabilistic model used to derive the probabilistic aspects of random processes. It's one of the important use is in the field of Natural Language Processing for Part of Speech Tagging [64], [65].

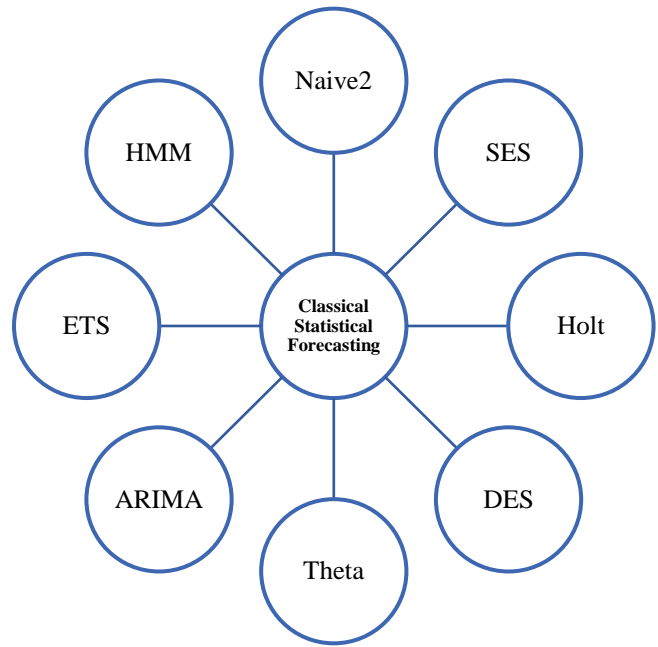

Fig. 9: Classical Statistical Forecasting Methods

The classical models can give a linear behavior between the target and the independent feature, whereas Machine Learning algorithms-based approaches can detect the non-linear behavior of the data set without knowing any other facts about the dataset. Moreover, these ETS or ARIMA models are the local model for each time-series model. In contrast, ML-based models learn jointly over the whole series. Hence, if a multivariate dataset, Machine Learning is a good choice over the classical methods [66] [67]. In other models like forecasting shoreline evolution for sandy coasts [68] or predicting exchange rates [69], impressive results have set the shift of the time series model from classical method to these Machine Learning-based methods.

### 4.4. Big Data and Machine Learning-Based Monitoring and Forecasting Systems

The data obtained from the real-time mechanism or downloaded needs to undergo a pre-processing architecture to get improved and reliable results. The data obtained from sources have some noise, missing values/attributes, or have errors. The process of cleaning the raw data is known as Data Pre-Processing or Data cleaning. After data is cleaned, data from all the sources is combined and stored; this huge data is managed by big data tools. After extracting all good quality data, target data mining is performed using machine learning algorithms.

In [11], Borges et al. collected the CO data for Sao Paulo's Metropolitan Area (SMA) by four IoT setups, and data was directed to Apache Hadoop through R studio. Using a popular HDFS (Hadoop Distributed File System) tool, MapReduce was implemented to process and analyze the data. Using Shiny, an R package that builds an interactive web application visualizes data. The data from all the sensor setups were compared, mean values per day for every sensor, sensor density, and summary of the measurements were potted using Shiny. It was found that the SMA's CO value was double the value established by WHO.

Ayele et al. [15] proposed real-time Monitoring and predicting system. Data was collected from the IoT setup, the collected data was stored on the web server, and the Long Short-Term Memory (LSTM) algorithm was used for prediction [70]. This experiment was performed by python 3.6.3 with TensorFlow as a backend. This experiment was concluded with good accuracy.

Moses et al. [12] collected data by IoT setup and suggested users with the alternative less polluted route on Google Map. Time series prediction of air quality was done by Neural Networks (NN) and Support Vector Machine (SVM) regression algorithm. For predicting AQI value, NN was used with sigmoid activation function, whereas SVM helps in error tolerance by individualizing hyperplanes.

Srivastava et al. [44] collected data via IoT setup and considered Central Pollution Control Board (CPCB) threshold values to send an emergency notification on the android application. Predictive analysis is done by two algorithms, Support Vector Regression (SVR) and Random Forest Regression (RFR). The models respective to these algorithms were conducted and evaluated using precision, recall, support, and F1 score. The accuracy in the SVR model was 90%, whereas in RFR it was around 99%. RFR performed well because many individual decision trees are involved in the algorithm, improving the overall result.

Wang et al. [71] proposed a data analysis and forecasting system for air pollution. The data used was taken from the Chinese air quality online monitoring and analysis platform using an API key. For storing atmospheric data, HDFS was used, while for calculation engine for industrial data, and Spark was used. Data were pre-processed to remove the duplicated or erroneous data. Data mining was done using Backward Propagation (BP) neural network. BP is a supervised learning algorithm, popularly used for prediction so that weights can be repeatedly adjusted and output is much closer to the expected vector. Data obtained through python crawler – AQI, CO, $NO_2$, $PM_{2.5}$, $PM_{10}$, city, date, wind speed, etc. was input to BP neural network, and it was found that wind speed value is one of the critical factors. The results were displayed using a visualization platform. This system used a combination of various technologies and proposed out an efficient method.

Kök et al. [30] have given a promising deep learning model for smart cities. Data used were collected from the CityPulse EU FP7 Project of Aarhus and Brasov cities in Denmark and Romania, respectively. Ozone and $NO_2$ pollutants were considered for this project. The dataset was divided into 69.5% and 30.5%, and python platforms Keras DL framework and TensorFlow were used. The data was trained on two algorithms, SVM and LSTM, for prediction. LSTM is a popular Recurrent Neural Network (RNN) that has feedback connections. Both models are compared by Root Mean Squared Error (RMSE) and Mean Absolute Error (MAE), followed by Confusion matrix for each model, precision, recall, and F1 score. The model with nearly good predictions for red alarms is considered yellow and green. Best precision and F1-Score as 98% and 97% respectively were given by LSTM, whereas for SVM, values were 95% (Precision) and 96% (F1 score) for red alarm. This system gave efficient and promising results for IoT data collected through CityPulse EU FP7 Project.

Nandini and Fathima [72] collected pollutant data and meteorological data from Central Pollution Control Boards (CPCB) and the Meteorological department. Then dataset was divided into validation data and test data as 90% and 10%, respectively. Then using K-Means Clustering, the data was converted into 3 clusters- Good, Moderate, and Unhealthy. After that, Multinominal Logistic Regression was used to observe the pattern between factors and results, and a decision tree was also used as a support tool to improve the conditional control statements. A confusion matrix is used to analyze the performance of both algorithms. The error rate value for the Regression model was found to be 0.428, and that of the Decision tree model was 0.666. Hence, the regression model was the best fit model.

In [22], Ameer et al. presented a comparative study for four advanced regression algorithms. The dataset consists of five cities of China – Shenyang, Beijing, Shanghai, Guangzhou, and Chengdu for a period from January 01, 2010, to December 31, 2015. Features of the dataset include $PM_{2.5}$ readings and other meteorological data of specific cities. By observing data, $PM_{2.5}$ had a negative correlation with other features, so data of every city was converted into correlation matrices. Prediction analysis was done using four advanced regression algorithms – GBR (Gradient Boosting Regression), DTR (Decision Tree Regression), MLP (Multilayer Perceptron Regression), and RFR (Random Forest Regression), and the results were evaluated by MAE, RMSE, and processing time. Decision Tree Regression is simple and took less processing time with MAE and RMSE, which is between 8 to 21 % and 0.06 to 0.24, respectively. MAE for Random Forest Regression ranged between 6 to 18 %, whereas RMSE ranged between 0.05 to 0.18, and processing time was lower than GBR and MLP. GBR had the highest error values. MLP was comparable to RFR. Out of all, RFR performed well after hyperparameter tuning using Spark. Hyperparameter tuning was done using a 10-fold cross-validation method and GridSearchCV function.

Mahalingam et al. [73] also studied the machine learning algorithms and their performance. Dataset was downloaded from Central Pollution Control Boards (CPCB) for New Delhi, India. Then whole data was divided into six categories – Good, Satisfactory, Moderate, Poor, Very Poor, and Severe. Neural Networks do prediction analysis with eight neurons in the hidden layer and six types of Support Vector Machines (SVM). Six SVMs with their accuracies are – Linear SVM (89.2%), Quadratic SVM (94.6%), Cubic SVM (94.6%), Fine Gaussian (62.2%), Medium Gaussian (97.3%) and Coarse Gaussian (78.4%). Neural networks provide an accuracy of 91.62%. Hence, Medium Gaussian is found the best fit.

In [31], Alaoui et al. proposed a model of air pollution using big data and ML algorithms. The data used is consists of $NO_2$ attributes ($NO_2$ units, $NO_2$ mean, $NO_2$ AQI, $NO_2$ $1^{st}$ max. value, and $NO_2$ $1^{st}$ max. hour) and meteorological data, which was taken from Kaggle. This data was pre-processed and was stored on Databricks. Then it was loaded and split into training and testing sets as 70% and 30%, respectively. Gradient-boosted trees (GBTs) as well as ML pipelines were used for data modeling and evaluated by Root Mean Squared Error (RMSE). Databricks were used for cloud-based data handling purposes. The obtained value of RMSE was 0.13. Hence the model is considered accurate.

In [14], Xiaojun et al. presented an IoT-based system. Environmental, as well as meteorological features, were taken into account. Two IoT setups were installed for other months – one was for January, February, November, December, and another for April, May, June, July, August, September, October. The ratio of Training Set: Validation Set: Test Set taken was 2:1:1. Data mining was done through neural networks (Input layer – 24, Hidden layer – 4, and Output layer – 1). Firstly, five meteorological factors were included, and 90% confidence was achieved using progressive regression) latter on input nodes were increased to 29 whereas hidden layer nodes to 6, and a study was done to compare the model's performances with and without meteorological factors. Another study was done by establishing artificial neural networks, and data of 5 years was taken as input. The model was studied based on the amount of data carried as input. Three years of data gave more close results than five years of data. It was concluded that meteorological data and the amount of data chosen as input data affects the results.

In [74], Ma et al. analyzed 171 features to investigate the factors that influence air quality. They have used non-linear ML algorithms MLR (Multiple Linear Regression), LR (Logistic Regression), Decision Tree-CART, kNN (K Nearest Neighbors), SVM (Support Vector Machine), ANN (Artificial Neural Network), BB (Bagging and Boosting), LR, BB SVM, RF (Random Forest), GBDT (Gradient Boosted Decision Trees), DNN (Deep Neural Network), and XGBoost (Extreme Gradient Boosting) to find the relationship between variables. In [75], Doreswamy et al. analyzed five different ML techniques, Decision Tree-CART, RFR (Random Forest Regressor), GBR (Gradient Boosting Regressor), KNR (K Neighbors Regressor), MLPR (Multilayer Perception Regression), to forecast the $PM_{2.5}$. They have used the TAQMN (Taiwan Air Quality Monitoring Network) dataset containing the data on air pollution in Taiwan from 2012 to 2017. The data are collected from 76 stations that are located at different places in the cities. The data is further pre-processed to fill the missing values. Further, metrological and air pollution data is fed to ML techniques to forecast the $PM_{2.5}$ level. The performance of the model is compared by using cross-validation and statistical metrics. They have found that GBR has outperformed as compared to other ML algorithms.

Xayasouk and Lee [76] developed an interesting model for $PM_{10}$ and $PM_{2.5}$ prediction. They took a dataset from January 01, 2017, to December 31, 2017, with 12 attributes – Wind Speed, Wind Direction, Temperature, Humidity, Rain, $PM_{10}$, $PM_{2.5}$, and Location information of cities Busan, Daegu, Daejeon, Gwangju, Incheon, Sejong, Seoul, and Ulsan. The model used for training and prediction purposes was the Stacked AutoEndorers (SAEs) Model. AutoEndorers are a type of neural network that reproduces its input and have one of each input, hidden, and output layer. If they are stacked in a hierarchical manner, they are known as SAEs. Data is trained in a greedy layer-wise manner. For each city, $PM_{10}$ and $PM_{2.5}$ were calculated, and RMSE was taken as performance metrics. Gwangju city performs best with 5.16 RMSE for $PM_{10}$ and 2.18 for $PM_{2.5}$. Results were even shown by plotting the actual and predicted values.

Barthwal et al. [77] presented an interesting solution. They worked on the data of Vasundhara Station, Ghaziabad, India, which was collected from the CPCB website. Data was consisted of both meteorological and air pollutant concentrations and taken at the hourly and daily average. The duration of data was from January 17, 2019 – January 30, 2020, which covers all seasons of that region. Meteorological data includes the attributes - Atmospheric Temperature (AT), Relative Humidity (RH), Solar Radiation (SR), Wind Speed (WS), Wind Direction (WD), and Atmospheric Pressure (AP) whereas air pollutant concentrations of $CO$, $O_3$, and $SO_2$. The missing values from the dataset were taken care of through Linear interpolation. The main issue discovered was the inconsistency in the correlation between $PM_{2.5}$ and $PM_{10}$. There was a strong correlation in the season of Monsoon – Winter, whereas, for Spring – Summer, it was poor. So, the forecasting models used were Multiple Linear Regression (MLR), Random Forests (RF), Support Vector Regression - Radial Basis Function (SVR-RBF), and Gradient Boosting Machine (GBM) due to non-linearity in the trend. Moreover, to improve the model performance and robustness, Hyperparameters were tuned for each algorithm. The ratio of Training Set and Testing Set was 80:20, and Stratified Random Sampling was used as cross-validation. The models were evaluated based on the – Metrics (coefficient of determination ($R^2$), Mean Error (ME), Absolute Mean Error (AME), Relative Error (RE) and Root Mean Square Error (RMSE)), Variation Importance Ranking (VIR) and Partial Plots (PPs). PPs for whole data and metrics for 7-days prediction for each season revealed GBM as the best model out of all. VIR analysis on the GBM algorithm showed CO as the most important feature for predicting PM levels.

Jeya et al. [78] uses Bi-directional LSTM (BiLSTM) model for $PM_{2.5}$ prediction. Data was extracted from UCI Machine Learning Repository, US for Beijing City and contains hourly data from January 01, 2010, to December 31, 2014, with features – Dew point, Temperature, Wind Direction, Wind Speed, Year, Month, Day, Hour, and $PM_{2.5}$. All data were normalized and then split into training and testing datasets in ratios 80 and 20, respectively. For correlation, a heatmap was plotted, and features with small correlations were removed for better accuracy. For training, BiLSTM was used with 32 neurons, batch size-16, epochs-20, Adam optimizer, and drop out is 0.2. The loss curve of the training set was plotted for the modeling of the test set accordingly. The Evaluation was done using RMSE, MAE, and Symmetric Mean Absolute Percentage Error (SMAPE). The RMSE of the proposed model was 9.86, whereas MAE and SMAPE were 7.53 and 0.1664, respectively, which was better than previously existing models.

Franceschi et al. [79] presented a prediction model of $PM_{2.5}$ and $PM_{10}$ based on Artificial Neural Networks (ANN) and k-means clustering for the city of Bogotá (Colombia). The data was collected from the Red de Monitoreo de Calidad del Aire de Bogotá (RMCAB) website for five years and from 13 stations. Missing data of the dataset can be considered up to the limit of 15% due to the device calibration, rain, or any other factor that could have caused the error or missing values in the collected data, and then data was normalized on the scale of [0, 1]. Principal Component Analysis (PCA) was used to reduce the dataset into less dimensional data while maintaining the main characteristics of the original dataset. Criteria of relevance were chosen as Cattell's Scree Test and featured with an eigenvalue of more than 0.2. The Back Propagation Neural Network (BPNN) model is considered as a forecasting model, whereas K means clustering groups the data before sending it to the forecasting model. X means an algorithm that is faster and starts with an assumption of the minimum number of clusters was chosen to find out the best number of clusters for each station. Quantitative analysis was done for measuring the performance, RMSE, MAE, and correlation coefficient (CC) values were 10.56, 14.01, and 0.72 respectively for $PM_{10}$, while for $PM_{2.5}$, it was 7.19, 9.34, and 0.66 respectively.

Chang et al. [80] recommended the system for $PM_{2.5}$ forecasting for Beijing, China. Data were collected from Taiwan Environmental Protection Agency (TEPA) and Central Weather Bureau (CWB) with 17 attributes for five years (Training Data) and 1-year data (Testing Data). Data pre-processing is done using Akima Normalization. The Aggregated LSTM (ALSTM) is used as a forecasting model with three input datasets (The local dataset, the neighboring station dataset, and industrial stations datasets) passed through two alternating LSTM and Dropout and finally to Dense, Merge, Dense, Output layer. Dropout layers are added to avoid overfitting of the model. This model was compared to LSTM, GBT, and SVR based on MSE and RMSE of 8 sequential future hours. ALSTM outclass other algorithms with minimum errors. Later the performance was also measured on a region-wise basis using MAPE, RMSE, and MAE.

Moursi et al. [81] proposed the Hybrid Non-linear AutoRegression with eXogenous input (NARX) model [82], which is mainly used for time series modeling. Dataset was taken from the University of California, Irvine website, which published the data from Beijing, China, for five years (2010-2014). The training segment took the first four years of the dataset, whereas the testing segment took the last year of dataset. Features used for this model were $PM_{2.5}$ value, cumulated wind speed and hours of rain. Hybrid NARX model was run on both a regular PC as well as on Raspberry Pi 4 but each time the model can take one ML model. The ML model taken into consideration were Long Short-Term Memory (LSTM), Random forests (RF), Extra Trees (ET), Gradient Boost (GB), Extreme Gradient Boost (XGB) and Random Forests in XGBoost (XGBRF). The evaluation of performance was done using Root Mean Square Error (RMSE), Coefficient of Determination ($R^2$), Index of Agreement (IA) and Normalized

Root Mean Square Error (NRMSE). The NARX-LSTM model worded out beat in terms of performance. But for accuracy and efficiency, NARX-XGBRF turned out best among all others.

Ketu et al. [83] propounded an algorithm to manage multiclass imbalance in the dataset using the Adjusting Kernel Scaling method (AKS), which is integrated with SVM classification. Dataset was taken from CPCB for Delhi region. In the algorithm, the kernel transformation function is calculated. This is done by calculating every support vector's weighting factor and parameter function at every iteration using Chi-square test [84]. The model was compared with already existing classification models, ADB (Ada Boost Algorithm), MLP (Multilayer Perceptron Algorithm), GNB (Gaussian NB Algorithm), Standard SVM (Support Vector Machine Algorithm) but out of all, the proposed model gave out best result with 99.66 accuracy. To avoid overfitting, K-Fold validation was used.

Zhong et al. [85] proposed a reinforcement learning based model named AirRL for the analysis of urban air quality. The authors claimed that their model is the first one to apply reinforcement learning for air pollution inference. The model has two modules: a station selector module that dynamically selects monitoring stations and a regressor module for air quality inference. Station selection is formulated using reinforcement learning to select optimal stations. The regressor module uses DNN to learn the relations among complex features. The authors recorded the data at 36 monitoring stations every hour during the period from 01/05/2014 to 30/04/2015 in Beijing, China. They used the dataset of only 30 stations which have fewer missing values for the analysis of $PM_{2.5}$ and $PM_{10}$. AirRL performed best with accuracies of 71.37% and 64.99% for $PM_{2.5}$ and $PM_{10}$ respectively against the other baseline methods considered for comparison. The RMSE values of AirRL were 35.7555 and 40.8122 for $PM_{2.5}$ and $PM_{10}$ respectively, which was better than the other baseline methods.

Table 6: Comparison of Different Systems with Big Data Technologies and Machine Learning Algorithms

| REF. | PROBLEM STATEMENT | REGION | PARAMETERS | DATA-SOURCE | TECHNIQUE | RESULTS |
|---|---|---|---|---|---|---|
| [11] | IoT and Big Data techniques-based study of CO | Sao Paulo, Brazil | CO | IoT setup (Sensor: MQ7) | Apache Hadoop and MapReduce through R studio. | Sao Paulo's CO readings were double what was established by WHO. |
| [15] | A monitoring and predicting system for air pollution | Vadodara, India | Temperature, humidity, and gas | IoT setup (Sensors: DHT11 and MQ135) | LSTM algorithm for prediction | The LSTM curve showed experiment accuracy. |
| [12] | Air pollution monitoring system based on IoT, and re-routing system based on ML | Coimbatore, India | Ozone, CO, $SO_2$, $NO_2$, and $PM_{2.5}$ | IoT setup (Sensors: MQ131 MQ7 110-602, $NO_2$ and $PM_{2.5}$ sensor) | Neural Network and SVR | Using this system Passenger was getting the re-routed less polluted way. |
| [44] | Study of ML algorithm to give improved IoT-based monitoring system | Rourkela, India | $PM_{2.5}$, $PM_{10}$ temperature, humidity, and CO | IoT setup (Sensors: HPMA115S0, SHT10) | SVR and RFR algorithm models | RFR gave better results. |
| [71] | An efficient air quality analysis system using advanced technologies | Beijing, China | AQI, CO, $NO_2$, $PM_{2.5}$, $PM_{10}$, city, date, wind speed | Chinese online Air quality monitoring platform. | HDFS, Spark MLlib for Back Propagation neural network | Cluster-based severity and pollutant vs. year were visualized. |

| Ref | Title | Location | Pollutants | Data Source | Methods | Results |
|---|---|---|---|---|---|---|
| [30] | A deep learning air quality predicting model | Ankara, Turkey | Nitrogen Dioxide | CityPulse EU FP7 Project | SVM and LSTM | With precision and F1-Score 98% and 97%, respectively, LSTM gave better results. |
| [72] | A better air quality predicting system using Machine learning Algorithms | India | SO, $NO_X$, HC, and Humidity | CPCB, Government of India | K-Means clustering to divide the data into 3 categories based on pollutant concentration then model is compared using Decision tree and Multinominal logistic regression. | Multinominal logistic regression performed better with an error rate of 0.428. |
| [22] | Comparative study of four advanced regression | Shenyang, Beijing, Shanghai, Guangzhou, and Chengdu (China) | Meteorological variables and PM2.5 | Heterogenous sources were used. Sensors collected data of $O_3$, $NO_2$, $SO_2$, and PM. | GBR, DTR, MLP, and RFR | RFR performed best out of all based on processing time, MAE. and RMSE with hyper-parameter tuning. |
| [73] | A promising Machine Learning Model for predicting AQI of New Delhi | New Delhi, India | $PM_{10}$, $PM_{2.5}$, $NO_2$, $SO_2$, CO, $O_3$, $NH_3$, Pb, and Meteorological variables. | CPCB, Government of India | Neural Networks and six kinds of SVMs | Results were evaluated based on accuracy: Medium Gaussian SVM got 97.3% whereas for Neural Network, 91.62%. |
| [31] | Air quality prediction model using IoT and big data analytics | Morocco | $NO_2$ | Kaggle | Databricks File System; GBTs as well as ML pipelines | RMSE is equal to 0.13; hence the model is accurate. |
| [14] | IoT-based Monitoring and predicting system | China | Smog, Inhalable particles, CO, Cl, HCl, and many other environmental and meteorological factors | IoT setup | Progressive regression neural networks | Meteorological inputs improved the prediction and the amount of data sent as input also affect the efficiency. |
| [74] | A multivariate analysis using ML model to investigate the influential factors of air quality | U. S. | Meteorological, Energy, Economical, Demographical, Transportation, and Environmental | GIS | MLR, LR, CART, KNN, SVM, ANN, BB LR, BB SVM, RF, GBDT, DNN, XGBoost | XGBoost performed better than others. |
| [75] | ML predictive model to forecast $PM_{2.5}$ concentration for air quality | Taiwan | Meteorological, time-series data | Air pollution stations | CART, RFR, GBR, KNR, MLPR | GBR performed better than others. |

| Ref | Title | Location | Data | Source | Model | Results |
|---|---|---|---|---|---|---|
| [76] | Deep Learning model for predicting $PM_{10}$ and $PM_{2.5}$. | South Korea | Fine dust particles ($PM_{10}$ and $PM_{2.5}$), Location Data, and Meteorological data. | Air Korean Website and Korean meteorological agency website | Stacked Auto Encoders | Performed best for Gwangju city with RMSE for $PM_{10}$ and $PM_{2.5}$ is 5.16 and 2.18 respectively. |
| [77] | Seasonality-based PM level prediction | Ghaziabad, India | 11 attributes including Meteorological and pollutant data | CPCB, Government of India | GBM, MLP, SVR-RBF, RF | GBM outperformed all and CO concentration affected the prediction of GBM the most. |
| [78] | Improved $PM_{2.5}$ predicting model | Beijing City, China | $PM_{2.5}$, Temperature, Dew point, WS, WD, and Timing data. | UCI Machine Learning Repository US | BiLSTM | The model performed better than existing models. The evaluation was done using RMSE, MAE, and SMAPE. |
| [79] | $PM_{2.5}$ and $PM_{10}$ forecasting model for Bogotá | Bogotá, Colombia | Wind Speed, Wind Direction Temperature, Precipitation, Relative Humidity, And Solar Radiation. | RMCAB | PCA, K Means, BPNN | The model was evaluated using RMSE, MAE, and CC. |
| [80] | A new model based on the LSTM method for predicting $PM_{2.5}$ | Taiwan | 17 attributes including local data, near station data, and industrial data. | TEPA and CWB | ALSTM | The model improved the accuracy with RMSE aspect 1.59, 0.44, and 2.77 and with MAE aspect 1.16, 0.91, and 5.27, better than existing models. |
| [81] | A hybrid NARX model for predicting $PM_{2.5}$ | Beijing, China | $PM_{2.5}$, cumulated wind speed and hours of rain | UCI Machine Learning Repository US | Hybrid NARX | The model was evaluated based on many metrics but NARX-LSTM and NARX-XGBRF were best. |
| [83] | Algorithm to improve multiclass imbalance in dataset | Delhi, India | $PM_{10}$, $SO_2$, $PM_{2.5}$, $O_3$, NOx, $NO_2$, NO, $NH_3$, CO, AQI, WD, $C_6H_6$, WS, RH, SR, BP, and AT | CPCB, Government of India | Scalable kernel-based SVM | The proposed model performed best with 99.66% accuracy. |
| [85] | AirRL: a reinforcement learning model for inference of urban air quality | Beijing, China | $PM_{2.5}$ and $PM_{10}$ | 36 Monitoring Stations | Reinforcement Learning and DNN | AirRL performed best with accuracies of 71.37% and 64.99% for $PM_{2.5}$ and $PM_{10}$ respectively. |

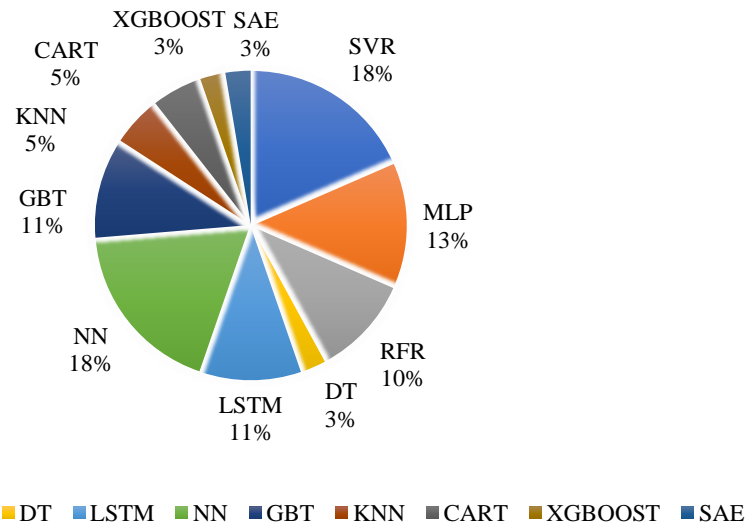

Fig. 10: Distribution of different algorithms used by various researchers.

Table 6 represents the various ML and Big Data models summarized, whereas Fig. 10 shows the distribution of multiple algorithms that gave better results to researchers. SVR is giving more stability to the predicting model than any other algorithm. NNs and LSTMs are emerging as a better solution for the problem because weights are assigned to each input feature, and nodes are set with the thresholds, which gives a better result than any other algorithm.

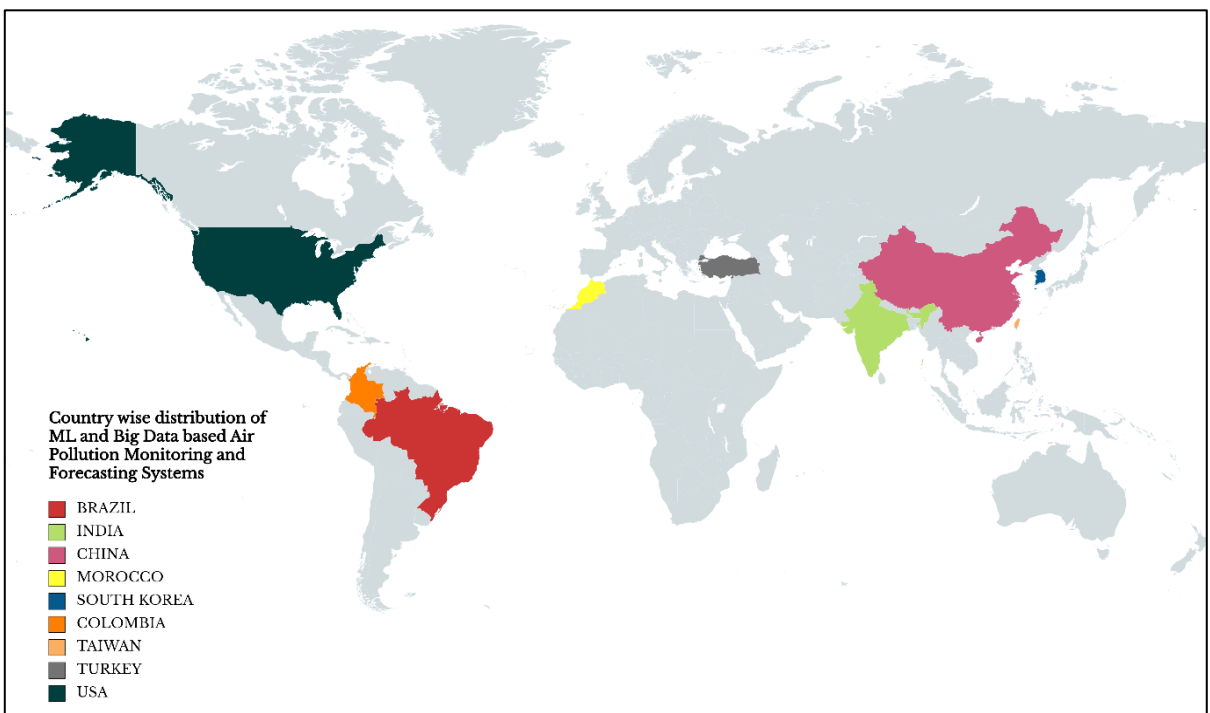

Fig. 11: Countries whose datasets are considered for air pollution models under study.

Further, Fig. 11 shows the countries whose datasets have been taken into account for predicting purposes. Table 7 distinguishes the above-discussed ML models in various aspects. i.e., is data collected with self-installed sensor setups, how missing data is handled, does the correlations were taken into account, and finally, hyperparameters were tuned or not.

Table 7: Comparing of models on various parameters.

| Ref. | Sensor setup | Handling missing data | Correlation between parameters | Hyperparameter tuning |
|---|---|---|---|---|
| [12] | ✓ | - | - | ✓ |
| [13] | ✓ | ✓ | ✓ | - |
| [15] | ✓ | - | - | - |
| [22] | ✓ | ✓ | ✓ | ✓ |
| [44] | ✓ | ✓ | - | ✓ |
| [71] | - | ✓ | - | ✓ |
| [30] | - | - | - | ✓ |
| [51] | - | ✓ | - | - |
| [73] | - | - | ✓ | - |
| [53] | - | ✓ | - | - |
| [74] | - | ✓ | ✓ | ✓ |
| [75] | - | ✓ | - | - |
| [76] | - | - | ✓ | - |
| [57] | - | ✓ | ✓ | ✓ |
| [58] | - | ✓ | ✓ | - |
| [79] | - | ✓ | ✓ | ✓ |
| [80] | - | ✓ | ✓ | - |
| [81] | ✓ | - | - | - |
| [83] | - | - | ✓ | ✓ |
| [85] | ✓ | - | ✓ | - |

## 5. RESEARCH ISSUES AND CHALLENGES

After examining and reviewing the literature in the above sections, some research issues and challenges for designing and implementing efficient models for air pollution monitoring and forecasting are underlined. Forecasting is a technique that is used consciously or unconsciously to predict what will happen and what is the likelihood of specific events. The research issues and challenges discussed below are also potential directions for future research.

*(i) Quality of Data:* The IoT infrastructure collects the data, but sometimes, due to poor network, sensor qualities, and connection faults, the data quality degrades. Good data quality with fewer missing and erroneous entries will give out the results with better accuracy.

*(ii) Quantity of Data:* In [73], Mahalingam et al. considered data records for one month (December 1 to 31) whereas Xiaojun et al. [14] highlight the amount of data considered is also an important factor for the accuracy of the model. The models must be tested on varying quantities of data. An appropriate amount of data must be considered for a better model.

*(iii) Real-Time Integrated Model:* A whole real-time integrated air quality monitoring and predicting system based on big data technologies and machine learning still lacks in the present scenario. Various factors affect the air quality of a particular area. To tackle all the dynamic changes, a stable and integrated model is still needed.

*(iv) Meteorological Factors:* Xiaojun et al. [14] also highlighted the importance of meteorological factors in such models. The accuracy of models could be improved by considering meteorological data of that specific area with pollutant data.

*(v) Uniformity of Sensor Setups:* Non-Uniform distribution of sensors across the cities affects the data quality. Sun et al. [13] underlined the importance of uniformity of sensor setups across the city for Monitoring and analyzing the city's air quality. To measure the air quality of a city, data collecting sensor setups must be kept at uniform distances.

*(vi) Number of Sensors Setups:* For collecting the air quality data, a good number of IoT setups must be installed across the region [13]. This will improve the overall stability of the model as well as data quality for mining purposes.

*(vii) Processing Time of Models:* The processing time of Machine learning models is also an important factor. In few proposed systems, machine learning models were solely evaluated on the accuracies and errors. An algorithm with high accuracy rates and a low processing time is preferred for an efficient air pollution forecasting and monitoring model.

*(viii) Number of Pollutants Taken into Account:* The air quality of the regions depends on the concentration of various toxic gases and particles. Some papers considered few gases as input for evaluating AQI. Ideally, all the gases causing air pollution should be taken into account. An efficient model is still required to consider all the intoxicants in the air.

*(ix) Checking Correlations:* ML models can be improved if the correlation between features can be determined before modeling it. The features with poor correlation with the target feature can be removed, resulting in improved performance of the model.

*(x) Hyperparameter Tuning:* Each data has its particular behavior; adjusting hyperparameters could help in optimizing the performance of the ML model.

## 6. CONCLUSION AND FUTURE WORK

Overexploitation of nature is reverting to earth's creatures in the form of deadly outcomes. The quality of air needs constant attention and evaluation. The advanced communication, computation, and analytics technologies, i.e., Internet of Things infrastructures, Big Data technologies, and Machine Learning algorithms, could provide us with more stable and efficient models for Monitoring and forecasting air pollution. This paper has reviewed and examined the recent air pollution monitoring and forecasting systems approach. Also, the existing model and systems have been compared on various parameters, and some significant research issues and challenges have been discussed in this paper. Further, some practical methods are suggested for improving the models. Presently, the air pollution sensing and monitoring tools and techniques suffer low efficiency, small range, and less accuracy. The real-time Monitoring and deployment of the available models are to be carried out as the future work of this study. The data-driven models can also be developed to predict, recommend, and monitor future work so that illnesses and climate change can be controlled. Finally, prevention and education can play a major role in controlling air pollution. Although, prevention and education will not reverse the adverse effects but will allow for sustainability for the future.

**APPENDIX: LIST OF ACRONYMS**

| Acronym | Full Name |
|---|---|
| ADB | Ada Boost Algorithm |
| ANN | Artificial Neural Network |
| AQI | Air Quality Index |
| ARIMA | Auto-Regressive Integrated Moving Average |
| HMM | Hidden Markov Model |
| BB | Bagging and Boosting |
| BiLSTM | Bi-directional LSTM |
| BMR | Bangkok Metropolitan Region |
| BP | Backward Propagation |
| BPNN | Back Propagation Neural Network |
| CO | Carbon Monoxide |
| COPD | Chronic Obstructive Pulmonary Disease |
| CPCB | Central Pollution Control Board |
| DES | Damped Exponential Smoothing |
| DNN | Deep Neural Network |
| DTR | Decision Tree Regression |
| EAQI | European Air Quality Index |
| EPA | Environmental Protection Agency |
| ETS | Error, Trend, and Seasonality |
| EU | European Union |
| GBDT | Gradient Boosted Decision Trees |
| GBM | Gradient Boosting Machine |
| GBR | Gradient Boosting Regression |
| GBR | Gradient Boosting Regressor |
| GBT | Gradient-Boosted Tree |
| GNB | Gaussian NB Algorithm |
| HAQ | Household Air Quality |
| HDFS | Hadoop Distributed File System |
| IAQI | Indoor Air Quality Index |
| IoT | Internet of Things |
| kNN | K Nearest Neighbors |
| KNR | K Neighbors Regressor |
| LR | Logistic Regression |
| LSTM | Long Short-Term Memory |
| MAE | Mean Absolute Error |
| ME | Mean Error |
| ML | Machine Learning |
| MLP | Multilayer Perceptron Algorithm |
| MLP | Multilayer Perceptron Regression |
| MLPR | Multilayer Perception Regression |
| MLR | Multiple Linear Regression |
| NAMP | National Air Monitoring Programs |
| NAQI | National Air Quality Index |
| NARX | Non-linear AutoRegression with eXogenous |
| NB-IoT | NarrowBand-Internet of Things |
| $NO_2$ | Nitrogen Dioxide () |
| NRMSE | Normalized Root Mean Square Error |
| $O_3$ | Ozone |
| PCA | Principal Component Analysis |
| PM | Particulate Matter or Particle Pollution |
| PPs | Partial Plots |
| RE | Relative Error |
| RF | Random Forest |
| RFR | Random Forest Regression |
| RMSE | Root Mean Squared Error |
| RNN | Recurrent Neural Network |
| SES | Simple Exponential Smoothing |
| SMAPE | Symmetric Mean Absolute Percentage Error |
| $SO_2$ | Sulphur Dioxide |
| SVM | Support Vector Machines |
| SVR | Support Vector Regression |
| SVR-RBF | Support Vector Regression - Radial Basis Function |
| TAQMN | Taiwan Air Quality Monitoring Network |
| TEPA | Taiwan Environmental Protection Agency |
| TVOC | Total Volatile Organic Compounds |
| US EPA | US Environmental Protection Agency |
| VIR | Variation Importance Ranking |
| WHO | World Health Organisation |
| XGB | Extreme Gradient Boost |
| XGBoost | Extreme Gradient Boosting |
| XGBRF | Random Forests in XGBoost |

## DECLARATIONS

**Authors' contributions:** Conceptualization: Amisha Gangwar, Sudhakar Sing, Richa Mishra, Shiv Prakash; Methodology: Amisha Gangwar, Sudhakar Sing, Richa Mishra, Shiv Prakash; Formal analysis and investigation: Amisha Gangwar, Sudhakar Singh, Richa Mishra, Shiv Prakash; Writing - original draft preparation: Amisha Gangwar, Sudhakar Singh; Writing - review and editing: Sudhakar Singh, Richa Mishra, Shiv Prakash; Resources: Amisha Gangwar, Sudhakar Singh, Richa Mishra, Shiv Prakash; Supervision: Sudhakar Singh, Richa Mishra, Shiv Prakash.

**Conflicts of interest:** The authors declare that they have no conflict of interest in this paper.

**Funding:** No funding was received for conducting this study.

**Availability of data and material:** Not applicable.

**Code availability:** Not applicable.